\documentclass[10pt,journal,compsoc]{IEEEtran}

\ifCLASSOPTIONcompsoc
  \usepackage[nocompress]{cite}
\else
  \usepackage{cite}
\fi

\ifCLASSINFOpdf
\else
\fi

\usepackage{dblfloatfix}
\usepackage{booktabs}
\usepackage{graphicx}

\usepackage{color}

\usepackage{xcolor}

\usepackage[hyphens]{url}
\usepackage[hidelinks, pagebackref=true,breaklinks=true,colorlinks,bookmarks=false]{hyperref}
\hypersetup{breaklinks=true}
\begin{document}
%
\title{VIPriors 1: Visual Inductive Priors for\\Data-Efficient Deep Learning Challenges}
%
%
%
%

\author{ Robert-Jan~Bruintjes,
        Attila~Lengyel,
        Marcos~Baptista~Rios,
        Osman~Semih~Kayhan,
        and~Jan~van~Gemert
\IEEEcompsocitemizethanks{
\IEEEcompsocthanksitem R. Bruintjes, A. Lengyel, O. S. Kayhan and J. van Gemert are with Delft University of Technology.\protect\\
E-mail: r.bruintjes@tudelft.nl
\IEEEcompsocthanksitem M. Baptista Rios is with University of Alcalá
}
}

%
%

\IEEEtitleabstractindextext{%
\begin{abstract}

We present the first edition of "VIPriors: Visual Inductive Priors for Data-Efficient Deep Learning" challenges. We offer four data-impaired challenges, where models are trained from scratch, and we reduce the number of training samples to a fraction of the full set. Furthermore, to encourage data efficient solutions, we prohibited the use of pre-trained models and other transfer learning techniques. The majority of top ranking solutions make heavy use of data augmentation, model ensembling, and novel and efficient network architectures to achieve significant performance increases compared to the provided baselines.
\end{abstract}

\begin{IEEEkeywords}
Visual inductive priors, challenge, image classification, object detection, instance segmentation, action recognition.
\end{IEEEkeywords}}

\maketitle

\IEEEdisplaynontitleabstractindextext

%
\IEEEpeerreviewmaketitle


%
%
%
%

\IEEEraisesectionheading{\section{Introduction}\label{sec:introduction}}
\IEEEPARstart{D}{ata} is fueling deep learning. Data is costly to gather and expensive to annotate. Training on massive datasets has a huge energy consumption adding to our carbon footprint. In addition, there are only a select few deep learning behemoths which have billions of data points and thousands of expensive deep learning hardware GPUs at their disposal. The Visual Inductive Priors for Data-Efficient Deep Learning workshop (VIPriors) aims beyond the few very large companies to the long tail of smaller companies and universities with smaller datasets and smaller hardware clusters. We focus on data efficiency through visual inductive priors.

At the first VIPriors workshop, hosted at ECCV 2020, we organized the first edition of the Visual Inductive Priors for Data-Efficient Deep Learning Challenges. Each of the four challenges invites competitors to solve a seminal computer vision task in a data-deficient setting. We challenge the competitors to submit solutions that can learn a good model of the dataset without access to the scale of data that powers state-of-the-art deep computer vision.

In this report, we discuss the outcomes of the first edition of these challenges. We discuss the setup of each challenge and the solutions that achieved the top rankings. We find that, in addition to typical methods used in deep learning challenges such as ensembling, the top competitors in all challenges heavily rely on data augmentation to make their solutions data-efficient. In our conclusion, we discuss the implications of this finding for the field of data-efficient deep learning.

\section{Challenges}


The Visual Inductive Priors for Data-Efficient Deep Learning Workshop accommodates four common computer vision challenges in which  the number of training samples are reduced to a small fraction of the full set:

\textbf{Image classification}: We use a subset of Imagenet~\cite{deng2009imagenet}. The subset contains 50 images from 1,000 classes for training, validation and testing.

\textbf{Object detection}: An MS COCO-2017~\cite{COCODetection2017} subset is used for this challenge. The subset includes approximately 6,000 training images.

\textbf{Semantic segmentation}:  A fraction of the Cityscapes dataset (MiniCity)~\cite{Cordts_2016_CVPR} is used for the segmentation challenge. The MiniCity dataset consists of a train, validation and test set of 200, 100 and 200 images, respectively.

\textbf{Action recognition}: The common UCF101 dataset~\cite{soomro2012ucf101} is used for the action recognition task. The training set consists of approximately 4.8k video clips.
 
We provide a toolkit\footnote{\url{https://github.com/VIPriors/vipriors-challenges-toolkit}} which consists of guidelines, baseline models and datasets for each challenge.
The competitions are hosted on the Codalab platform. Each participating team submits their predictions computed over a test set of samples for which labels are withheld from competitors.

The challenges include certain rules to follow:
\begin{itemize}
    \item Models ought to train from scratch with only the given dataset.
    \item The usage of other data rather than the provided training data, pretraining the models and transfer learning methods are prohibited. 
    \item The participating teams need to write a technical report about their methodology and experiments.
\end{itemize}

\subsection{Classification}

Image classification is the most common problem in computer vision and the state-of-the-art classification methods are nourished by the usage of massive datasets, such as the private JFT-300M~\cite{sun2017revisiting}, and computational power to train the models. In addition, for domains like medical imaging, the amount of labeled data is limited and the collection and annotation of such data relies on domain expertise. Therefore, the design of data efficient methods is crucial. 

In our image classification challenge we provide a subset~\cite{kayhan2020translation} of the Imagenet dataset~\cite{deng2009imagenet} consisting of 50 images per class for each of the train, validation and test splits. 
Participants can only access the labels of the train and validation splits. We provide two baselines based on the Resnet-50~\cite{he2015deep} architectures, one with "same" padding and one with "valid" padding~\cite{kayhan2020translation}. 
The classification challenge had seven participating teams, of which five teams submitted a report. The final ranking and the results can be seen in Table~\ref{tab:classification}.
The method of Sun et al.~\cite{sun2020visual} obtains $73\%$ accuracy and holds the first place of the challenge. Luo et al.~\cite{luo2020technical} and Zhao et al.~\cite{zhao2020distilling} share the second place by $70\%$ and $69\%$, respectively. Kim et al.~\cite{kim2020dataefficient} achieve third place with $67\%$ of accuracy.
All the winning teams utilize strong data augmentation and backbone architectures to train their models. To obtain the final version of their models, the teams use validation set for training and combine multiple models by using model ensemble techniques.

\begin{table}[h]
\centering
\caption{Final rankings of the Image Classification challenge.}
\renewcommand{\arraystretch}{1.3}
\begin{tabular}{@{}llc@{}}
\toprule
Ranking & Teams                & Top-1 Accuracy \\ \midrule
1       & \textbf{Sun et al.}~\cite{sun2020visual}   & \textbf{0.73}           \\
2       & Luo et al.~\cite{luo2020technical}   & 0.70           \\
2       & Zhao et al.~\cite{zhao2020distilling}  & 0.69           \\
3       & Kim et al.~\cite{kim2020dataefficient}  & 0.67           \\ 
4       & Liu et al.~\cite{Liu2020DiversificationIA}  &
0.66            \\
\bottomrule
\end{tabular}
\label{tab:classification}
\end{table}
\subsubsection{First place}
\begin{figure}[h]
	\centering
	\includegraphics[width=\linewidth]{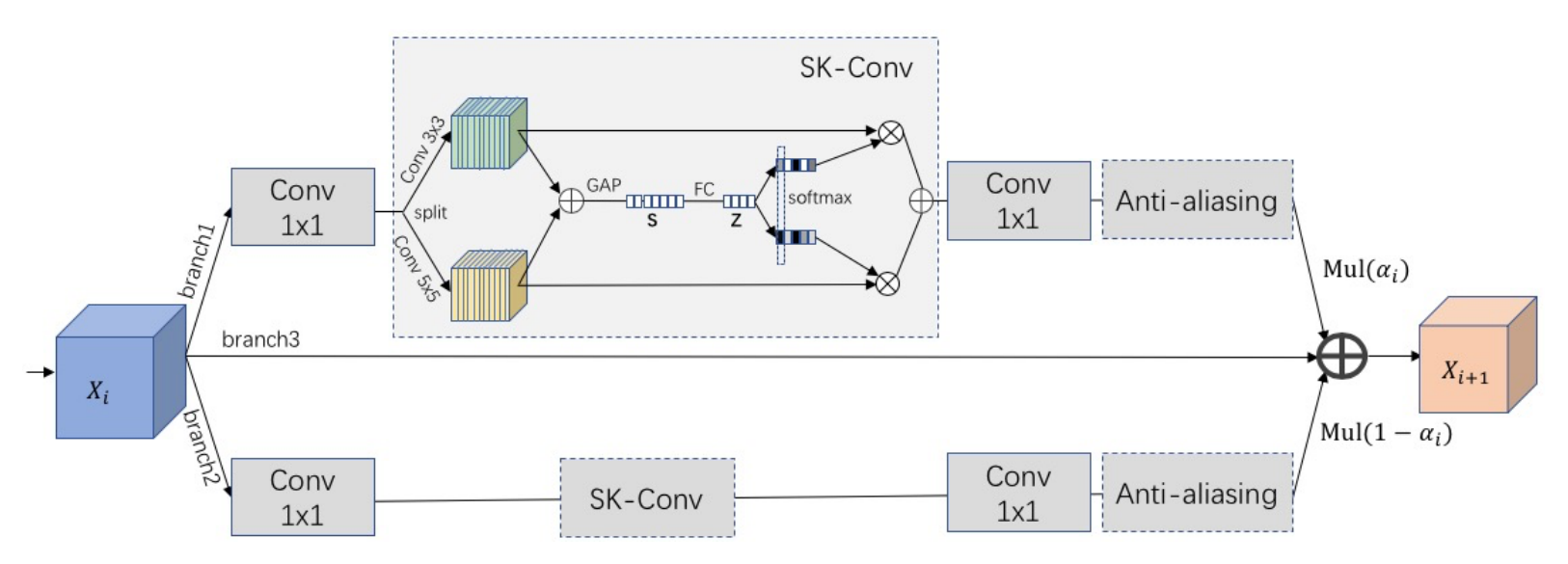}
	\caption{Dual Selective Kernel network~\cite{sun2020visual}. The block consists of 2 Selective Kernel convolution and 1 skip connection.}
	\label{fig:dsknet}
\end{figure}

Sun et al.~\cite{sun2020visual} propose Dual Selective Kernel network (DSK-net) to achieve data-efficient learning (Fig.~\ref{fig:dsknet}). The DSK-net block has 3 branches, 2 Selective Kernels (SK)~\cite{li2019selective} and 1 skip connection. After each SK, an anti-aliasing~\cite{zhang2019shiftinvar} block is attached. The team also employs 3 loss functions: positive class loss, center loss~\cite{wen2016discriminative} and tree supervision loss inspired by~\cite{wan2020nbdt}. Cutmix data augmentation~\cite{yun2019cutmix} is applied during training. In the end, a model ensemble with 16 models obtains 73.03\% accuracy.

\subsubsection{Second place}
The teams \cite{luo2020technical} and \cite{zhao2020distilling} share the second place due to a logistic issue about technical reports.

Luo et al.~\cite{luo2020technical} use several strong backbones, loss functions, data augmentation and model ensembling. The team trains ResNest-101~\cite{zhang2020resnest}, TresNet-XL~\cite{ridnik2021tresnet} and SEResNeXt-101~\cite{hu2018squeeze} with only cross entropy and combination of cross entropy - triplet loss~\cite{hermans2017defense}, and cross entropy - ArcFace loss~\cite{deng2019arcface} functions. Besides, label smoothing~\cite{szegedy2016rethinking} is applied to regularize the classifier layer. AutoAugment~\cite{cubuk2018autoaugment} with 24 different policies and Cutmix~\cite{yun2019cutmix} data augmentation methods are utilized to mitigate the overfitting problems. The final method attains $70\%$ top-1 accuracy.

\begin{figure}[h]
	\centering
	\includegraphics[width=\linewidth]{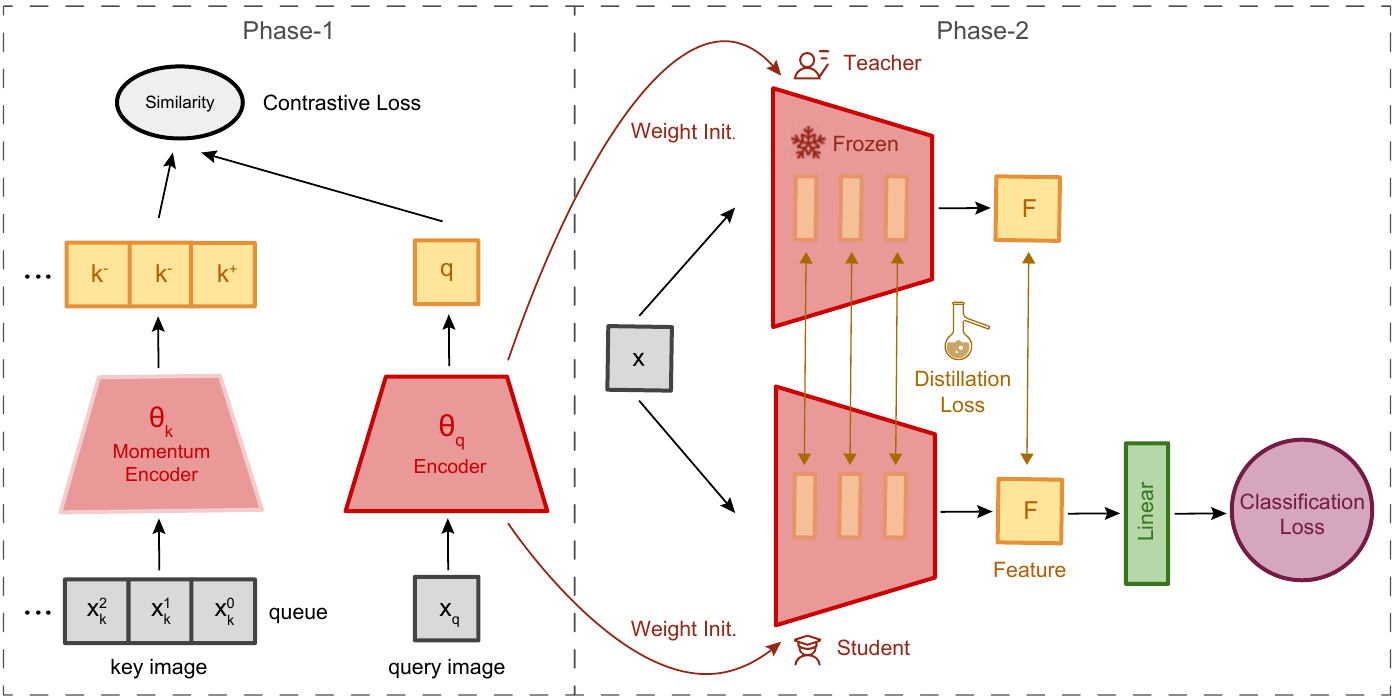}
	\caption{In Phase-1, the backbone is trained by using MOCO~\cite{He_2020_CVPR} contrastive learning method. In Phase-2, the trained backbone is employed for knowledge distillation to train student network.}
	\label{fig:distill}
\end{figure}

Zhao et al.~\cite{zhao2020distilling} propose a method which consists of two stages: (i) A teacher network is trained with contrastive learning, MOCO v2~\cite{He_2020_CVPR}, to obtain a feature representation and (ii) the knowledge of the teacher network is transferred to student network by knowledge distillation~\cite{heo2019comprehensive}. Meanwhile, the student network is also finetuned with labels (Fig.~\ref{fig:distill}). To obtain the final result, they increase the input size as 448x448 and use ResNeXt101~\cite{xie2017aggregated}, AutoAugment~\cite{cubuk2018autoaugment}, label smoothing~\cite{muller2019does}, ten crops and model ensembling. The method reaches $69\%$ accuracy.

\subsubsection{Third place}

\begin{figure}[h]
	\centering
	\includegraphics[width=0.9\linewidth]{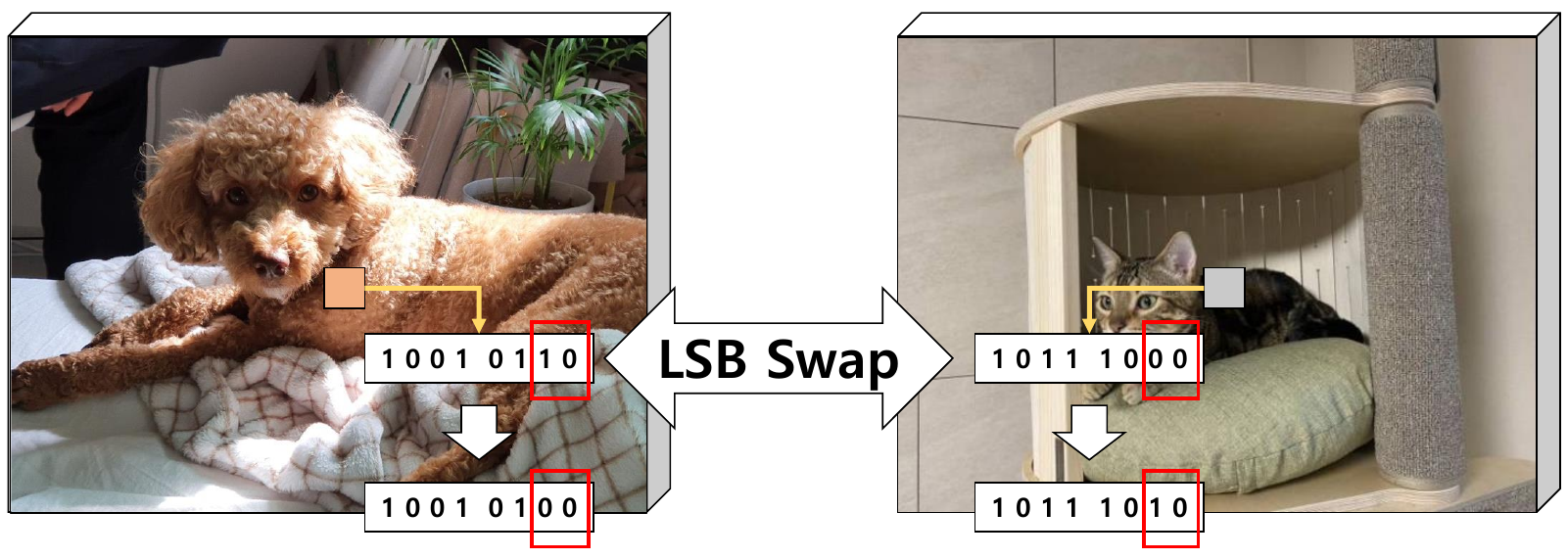}
	\caption{Swapping Low Significant Bit (LSB) of all image pixels with different image to augment the images. Last 2 LSBs of cat and dog images are exchanged for every equivalent pixel location.}
	\label{fig:lsb}
\end{figure}

Kim et al.~\cite{kim2020dataefficient} focus on data augmentation, loss function and ensemble methods to obtain good classification performance. They train EfficientNet~\cite{tan2019efficientnet} backbones and Low Significant Bit (LSB) swapping between image pixels is used as a data augmentation method (Fig.~\ref{fig:lsb}) in addition to RandAugment~\cite{cubuk2020randaugment}. They propose focal cosine loss, which is a combination of the focal~\cite{lin2017focal} and cosine~\cite{barz2020deep} losses. Besides,  Exponential Moving Average, dropout and drop connection are employed during training. Plurality voting ensemble method with 10 models and Test Time Augmentation are applied to obtain the final result as $67\%$.
\subsection{Object Detection}

\begin{table*}[!t]
\centering
\caption{Final rankings of the Object Detection challenge.}
\renewcommand{\arraystretch}{1.3}
\begin{tabular}{ll|cccccc}
\hline
  &                                                                        & \textbf{AP}             & AP @ 0.50      & AP @ 0.75      & AP (S)     & AP (M)    & AP (L)     \\ \hline \hline
1. & \textbf{Shen et al.} \cite{Shen2020} & \textbf{0.394} & \textbf{0.563} & \textbf{0.427} & 0.162          & \textbf{0.435} & \textbf{0.611} \\
2. & Gu et al. \cite{gu20202nd}                            & 0.366          & 0.529          & 0.400          & 0.224          & 0.387          & 0.456          \\
3. & Luo et al. \cite{luo2020vipriors}                     & 0.351          & 0.530          & 0.392          & \textbf{0.250} & 0.393          & 0.415          \\
  & \textit{baseline}                                                      & 0.049          & 0.112          & 0.036          & 0.010          & 0.038          & 0.089          \\
  \hline
\end{tabular}
\label{tab:detection}
\end{table*}


Object detection is a crucial part of many applications of computer vision, from object picking robot arms to self-driving cars looking out for other cars and pedestrians. Labeling for object detection is a time-consuming effort, creating a need for data-efficient object detectors. This challenge aims to stimulate developing such solutions.

The training data for this challenge is a subset of 5,873 samples from the Microsoft COCO dataset~\cite{lin2015microsoft}. Evaluation is performed by computing Average Precision @ 0.50:0.95 over the MS COCO validation dataset, matching the evaluation protocol of the COCO 2017 Object Detection Task~\cite{COCODetection2017}.

The baseline solution is a Faster R-CNN~\cite{ren2016faster} model with ResNet-18~\cite{he2015deep} FPN~\cite{lin2017feature} backbone, trained from scratch for 51 epochs with initial learning rate 0.02 and decay at epoch 48.

\subsubsection{Final rankings}

Seven teams submitted solutions to the evaluation server, of which three teams submitted a report to qualify their submission to the challenge. The final rankings are shown in Table \ref{tab:detection}.

\subsubsection{First place}
Shen et al. \cite{Shen2020} achieve the winning solution by combining a number of different approaches. The architecture is a two-stage Cascade-RCNN~\cite{cai2018cascade} with multiple backbones. The authors infuse global context features into each ROI feature to help the network deal with extreme circumstances such as multiple targets and occlusions. Instead of Non-Maximum Suppression the authors use Weighted Boxes Fusion~\cite{Solovyev_2021} to fuse the predictions of the different backbone networks. Other small changes to the architecture are detailed in the report~\cite{Shen2020}. The network is trained with a combination of different data augmentation techniques: a novel method called "bbox-jitter", which randomly translates bounding boxes slightly, as well as the established methods Grid-mask~\cite{chen2020gridmask} and Mix-up \cite{zhang2018mixup}.

\subsubsection{Second place}
Gu et al.~\cite{gu20202nd} start building their solution with a Cascade R-CNN~\cite{cai2018cascade} with R50-FPN-DCNv2 backbone. The authors add many different methods to their solution, including several modules from the Libra R-CNN~\cite{pang2019libra}, Guided Anchoring~\cite{wang2019region} and Generalized Attention~\cite{zhu2019empirical} frameworks, the ResNet-D variation on ResNets~\cite{he2018bag}, and TSD~\cite{song2020revisiting}. The Albumentations library~\cite{Buslaev_2020}, the AutoAugment policies~\cite{zoph2019learning} and the Stitchers method~\cite{chen2020stitcher} are used for data augmentation, as well as a novel variation on the Mosaics method~\cite{bochkovskiy2020yolov4}, called Mosaics-SC, which couples objects from the same supercategory. The final model is an ensemble of different ResNets sharing the described method.

\subsubsection{Third place}
Luo et al.~\cite{luo2020vipriors} use a Scratch Mask R-CNN~\cite{zhu2019scratchdet} architecture with a ResNet-101~\cite{he2015deep} backbone, Soft-NMS~\cite{bodla2017soft} and a custom weighting of the components of the loss function to put more emphasis on the classification loss. Data augmentation is used to partially decrease the class imbalance, and thirty different augmentations from the Albumentations library~\cite{Buslaev_2020} are applied to all images in training.

\subsection{Segmentation}
Semantic segmentation and scene understanding is a key problem in computer vision with applications ranging from autonomous driving, medical imaging and remote sensing. Our challenge is based on the popular Cityscapes~\cite{Cordts_2016_CVPR} dataset of street images and we provide reduced training, validation and test sets of 200, 100 and 200 samples, respectively. The pixel-level segmentation predictions are evaluated by the mean Intersection over Union (mIoU) metric. Our baseline method is based on the UNet architecture~\cite{Ronneberger2015Medical} with added Batch Normalization~\cite{IoffeS15} layers.

\subsubsection{Final rankings}
Ten teams submitted solutions to the evaluation server, of which five teams submitted a report to qualify their submission to the challenge. The final rankings are shown in Table \ref{tab:segmentation}.

\begin{table}[!t]
\centering
\caption{Final rankings of the Object Segmentation challenge.}
\renewcommand{\arraystretch}{1.3}
\begin{tabular}{ll|cc}
\hline
  &                                                  & mIoU                 & Accuracy             \\
  \hline \hline
1 & \textbf{Weitao et al.~\cite{Mj_2020}} & \textbf{65.64}   & \textbf{83.01}       \\
2 & Liu et al.~\cite{Liu2020}         & 65.61                & 83.11                \\
3 & Hsu et al.~\cite{Hsu_Ma_2020}         & 64.35                & 83.03                \\
4 & Yeşilkaynak et al.~\cite{Yesilkaynak_Sahin_Unal_2020} & 58.03                & 81.68                \\
5 & Pytel et al.~\cite{Pytel2020}       & 43.06                & 78.85                \\
  & \textit{baseline}          & 38.77                & 78.1                 \\ \hline
\end{tabular}
\label{tab:segmentation}
\end{table}

\subsubsection{First place}
Weitao et al.\cite{Mj_2020} propose a multi-scale version of CutMix~\cite{Yun_2019_ICCV} to boost the occurrence of scarce data classes through data augmentation. The method first trains a segmentation model and samples random crops from the training data from the worst performing classes. Then CutMix is used to augment the training data using the sampled crops and finally the model is retrained. The method uses the HRNet~\cite{Wang2019Deep} architecture with additional scale attention.

\subsubsection{Second place}
Liu et al.~\cite{Liu2020} propose to diversify the models, the data and the test samples by using a combination of extensive data augmentation and model ensembling. Firstly, the method extends the Augmix~\cite{hendrycks2020augmix} algorithm to semantic segmentation. This is done by maintaining a pool of photometric data augmentation methods, i.e. augmentations that do not alter the spatial characteristics of an image, and transform an image through multiple branches of different combinations of augmentations. The branches are then combined both with each other and with the original input image as a weighted sum. Second, the method combines the HRNetv2~\cite{Wang2019Deep} architecture, Object Contextual Representations~\cite{Yuan2019Object}, online test time augmentations and Online Hard Example Mining~\cite{Shrivastava_2016_CVPR}. Finally, multiple models are trained and are combined as a Frequency Weighted ensemble such that low frequency classes have higher weights.

\subsubsection{Third place}
Hsu et al.~\cite{Hsu_Ma_2020} introduces an edge-preserving loss to force the edge maps of the predictions and ground-truth labels to overlap. The edge maps are extracted using simple Sobel filtering and hard-thresholding. Additionally, the pixel occurrence of rare classes is increased by pasting augmented crops from these classes in the training images. Furthermore, the HANet~\cite{choi2020cars} architecture is improved by changing the feature extractor network to ResNeSt~\cite{zhang2020resnest}.

\subsection{Action Recognition}
Many of the most popular Action Recognition models consist of very deep networks whose training process requires a massive amount of data, in the form of frames or clips. This fact becomes one of the main obstacles on occasions when, for example, there is not enough data available or resources are insufficient to be able to adjust the model correctly.

With this challenge, we want to encourage Action Recognition researchers to develop efficient models capable of extracting visual prior knowledge from data. To this aim, we have adapted the UCF101~\cite{soomro2012ucf101} dataset. From the official splits, we have built a reduced version of the training and test sets with 4.8K and 3.8K videos, respectively. Additionally, we also provide a validation set with 4.7K videos. 

As metric, we evaluate the average classification accuracy over all classes on the test set. The accuracy for one class is defined as $ \mathrm{Acc} = \frac{P}{N} $, where P corresponds to the number of correct predictions for the class being evaluated and N to the total number of samples of the class. The average accuracy is the average of accuracies over all classes.

\subsubsection{Final Rankings}
Ten teams submitted solutions to the evaluation server, of which four teams submitted a report to qualify their submission to the challenge. The final rankings are shown in Table \ref{tab:actionrecognition}.

\begin{table}[h]
\centering
\caption{Final rankings of the Action Recognition challenge.}
\renewcommand{\arraystretch}{1.3}
\begin{tabular}{@{}llc@{}}
\toprule
Ranking & Teams                                 & Top-1 Accuracy (\%) \\ \midrule
1       & \textbf{Dave et al.} \cite{Dave2020vipriorsar} & \textbf{90.83}           \\
2       & Chen et al. \cite{Chen2020vipriorsar} & 88.30           \\
3       & Luo et al. \cite{Luo2020vipriorsar}   & 87.60           \\ 
4       & Kim et al. \cite{Kim2020vipriorsar}   & 86.04            \\
\bottomrule
\end{tabular}
\label{tab:actionrecognition}
\end{table}

\subsubsection{First Place}
Dave et al.~\cite{Dave2020vipriorsar} assume that motion between frames serves as a good prior. For their submission, they propose a two-stream configuration (RGB + Flow) in which each stream combines some of the best models for video processing. While the RGB stream ensembles four popular networks: I3D~\cite{Carreira2017i3d}, C3D~\cite{Tran2015c3d}, R3D and R2+1D~\cite{Tran2018r21d}, the Flow stream combines two I3D~\cite{Carreira2017i3d} with different input clip length. Finally, results of the two streams are fused. During training, authors also apply different data augmentation strategies such as random crops, horizontal flipping or frame skipping.

\subsubsection{Second Place}
Chen et al.~\cite{Chen2020vipriorsar} achieve the second place by using a two-stream architecture (RGB + Flow) based on a modified C3D~\cite{Tran2015c3d} network. The enhanced C3D model adds a new term called Temporal Central Difference Convolution (TCDC) to the computation of the vanilla 3D convolution which helps including information from the adjacent components of the 3D receptive field. Authors also suggest that RGB might contain lots of noisy details. Therefore they utilize a Rank Pooling~\cite{Ferando2016rp} representation for the RGB stream.

\subsubsection{Third Place}
Since common video analysis configurations for action recognition are based on two-stream architectures (RGB + Flow) in which the extraction of the optical flow is computationally intensive, Luo et al.~\cite{luo2020vipriors} decided to takle the challenge using a more efficient setup: SlowFast \cite{Feichtenhofer2019sf} model. To improve results while remaining efficient, they also fuse the SlowFast with TSM \cite{Lin2019tsm} modules. In addition to this setup, the authors applied several data augmentation techniques: center/random crop, horizontal flip and normal/reverse video reproduction.

\section{Conclusion}
The first VIPriors challenges have provided a valuable insight in the current state-of-the-art methods and techniques in practical, low-data computer vision. In all challenges the contenders were able to achieve surprisingly competitive performance compared to models trained on the full training set. Heavy data augmentation~\cite{bochkovskiy2020yolov4,chen2020gridmask,chen2020stitcher,yun2019cutmix,zhang2018mixup,zoph2019learning} proved to be extremely effective, both in expanding the training set by synthetically generating new samples as well as in improving the class balance by resampling infrequently occurring classes. Given the increasing popularity of self-supervision and contrastive learning~\cite{chen2020simple}, we believe data augmentation will continue to play a significant role in deep learning. Other effective methods included model ensembling, the use of novel and efficient network architectures~\cite{tan2019efficientnet,Wang2019Deep,ridnik2021tresnet,hu2018squeeze}, and techniques such as test-time augmentation. Interestingly, but perhaps not surprisingly, most high-ranking submissions were based on well-established methods, whereas novel ideas were often lagging behind in terms of performance. Nevertheless, we encourage novel ideas, as they might be the breakthrough solutions of the future.

\Urlmuskip=0mu plus 1mu\relax
\bibliographystyle{splncs04}
\bibliography{egbib}

%




\end{document}